\pdfoutput=1
\PassOptionsToPackage{usenames,dvipsnames}{xcolor}
\documentclass[11pt]{article}

\usepackage[preprint]{acl}

\usepackage{times}
\usepackage{latexsym}
\usepackage{bm}
\usepackage{marvosym}
\usepackage{amsmath,amssymb}%
\usepackage{amsthm}%
\usepackage{xcolor}%
\usepackage{graphicx}%
\usepackage{url}
\usepackage[many]{tcolorbox}
\usepackage{enumitem}
\usepackage{setspace}
\usepackage{caption}
\usepackage{wrapfig}
\usepackage[caption=false, font=footnotesize, captionskip=0pt]{subfig}
\usepackage{array}
\usepackage{multirow}
\usepackage{booktabs}
\usepackage{soul}
\usepackage[many]{tcolorbox}
\usepackage{subfig}

\usepackage{algorithm}
\usepackage[noend]{algpseudocode}
\algrenewcommand\algorithmicfunction{}

\usepackage[T1]{fontenc}
\usepackage[utf8]{inputenc}

\usepackage{microtype}

\usepackage{inconsolata}

\usepackage{graphicx}

\theoremstyle{definition}
\newtheorem{defn}{Definition}[section]

\theoremstyle{definition}

\theoremstyle{remark}

\title{Measuring temporal effects of agent knowledge by date-controlled tool use}

\author{
 \textbf{R. Patrick Xian\textsuperscript{1}},
 \textbf{Qiming Cui\textsuperscript{2,1}},
 \textbf{Stefan Bauer\textsuperscript{3}},
 \textbf{Reza Abbasi-Asl\textsuperscript{1}},
\vspace{5pt}
\\
 \textsuperscript{1}UC San Francisco,
 \textsuperscript{2}UC Berkeley,
 \textsuperscript{3}Technical University of Munich \& Helmholtz AI
\\
 \normalsize{
   \Letter: xrpatrick@gmail.com, qcui@berkeley.edu, st.bauer@tum.de, reza.abbasiasl@ucsf.edu
 }
}

\begin{document}
\maketitle
\begin{abstract}
Temporal progression is an integral part of knowledge accumulation and update. Web search is frequently adopted as grounding for agent knowledge, yet an improper configuration affects the quality of the agent's responses. Here, we assess the agent behavior using distinct date-controlled tools (DCTs) as stress test to measure the knowledge variability of large language model (LLM) agents. We demonstrate the temporal effects of an LLM agent as a writing assistant, which uses web search to complete scientific publication abstracts. We show that the temporality of search engine translates into tool-dependent agent performance but can be alleviated with base model choice and explicit reasoning instructions such as chain-of-thought prompting. Our results indicate that agent design and evaluations should take a dynamical view and implement measures to account for the temporal influence of external resources to ensure reliability\footnote{The code and datasets for the work are available at \url{https://github.com/RealPolitiX/agent_oost}.}.
\end{abstract}

\section{Introduction}
\label{sec:intro}
\begin{figure*}[ht]
% 0.53, 0.42
    \centering
    \subfloat[]{
        \includegraphics[width=0.51\linewidth]{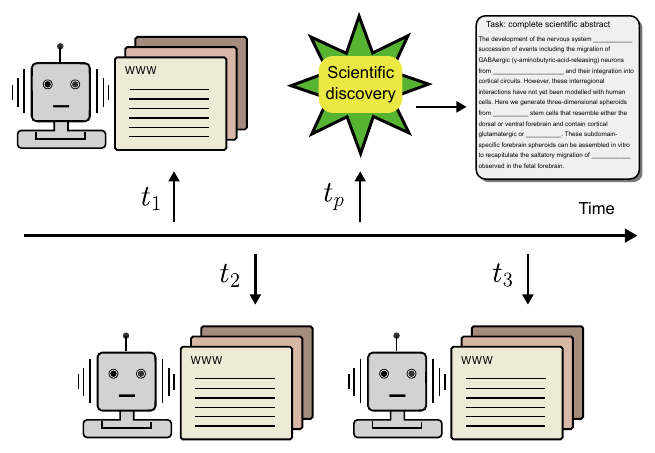}
    }
    \hfill
    \subfloat[]{
        \includegraphics[width=0.41\linewidth]{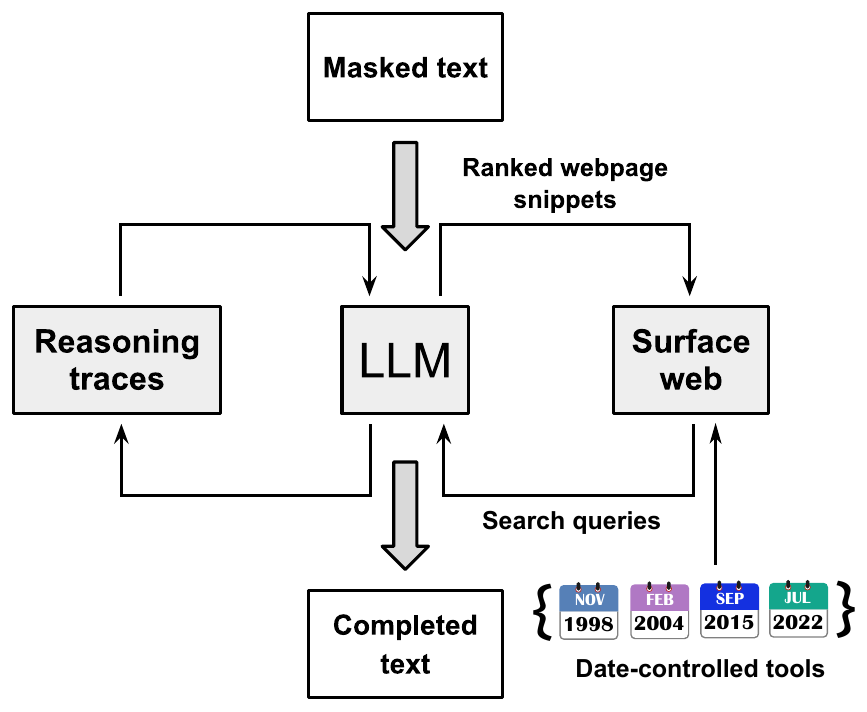}
    }
    \caption{(a) Illustration of the stress testing framework for agent knowledge, $t_p$ indicates the time of publication (b) Temporal tool selection in a ReAct-style agent that performs text completion task in (a) with a selected tool.}
    \label{fig:schematic}
\vspace{-1em}
\end{figure*}

AI agents based on LLMs and equipped with tools \citep{mialon_augmented_2023,wang_tools_2024} are well-suited for complex real-world tasks \citep{gao_empowering_2024,theagentcompany_2024} because of their extended capabilities. Their potential to become virtual assistants, paraprofessionals, or ``copilots" holds promise for improving the productivity and creativity of the scientific, medical workforces and beyond \citep{wachter_genai_2024,wornow_automating_2024,bousetouane_agentic_2025}. The evaluation standards for AI agents are still in flux \citep{kapoor_agent_matters_2024,hojmark_analyzing_2024} and they are urgently needed in specialized domains and realistic scenarios where the the outcomes convey greater bearing on their adoption. Recent works demonstrated the feasibility of LLMs in predicting temporal events \citep{ye_mirai_2024} and carry out time series forecasting \citep{tang_time_2024}, but their equivalents in agentic systems are not yet realized. Scientific knowledge and claims have a strong temporal dependence but they have so far been less studied in the context of generative language models \citep{zhao_stc_2024,chroknowledge_2024}. We devised a text completion task as a proxy to measure the agent's usability as a writing assistant with access to external sources (see Fig. \ref{fig:schematic}a).

% We investigate the temporal bias imposed by the tool on the agent performance.
% from the tool can translate into positional bias in the context window according to the information the agent receives, while the cognitive bias manifests itself in the base model of the agent, potentially manipulating its decision-making and performance \citep{carroll_manipulation_2023}.
Web search is an essential tool for grounding agent knowledge to the current and bygone worlds \citep{pavlick_symbols_2023} and it appears in many applications as a capability extender for models \citep{webarena_2024,song_beyondapi_2024}. Nevertheless, web search is subject to the recency and primacy bias of the search engine \citep{lawrence_searching_1998} and the cognitive bias of the users who seek and collect information \citep{lau_people_2007}. The term \textit{search engine manipulation effect} \citep{epstein_search_2015} was coined to refer to the search results' influence on public opinions of societal issues. Independent of search engines, factual and scientific knowledge also experience constant but necessary updates over time \citep{arbesman_half-life_2013}.

While temporal generalization remains challenging for language models \citep{lazaridou_mind_2021,wallat_temporal_2024}, explicit tuning of time-related tool parameters in LLM agents can offer an alternative way to reduce \textit{temporal effects} of the base model. These effects are a source of performance reliability issues of agentic systems that warrant investigation \citep{ye_toolsword_2024}. Date control in reality can manifest \textit{passively} because the tool-interfaced computer programs have an intrinsic time stamp or a versioned release over time \citep{zhang_tempsearch_2009}. Alternatively, date control can be imposed \textit{actively} because of copyright, paywall, or local policy. Content access in the past can be controlled retroactively as policy changes \citep{aral_digital_2021}. From a technical standpoint, invoking different DCTs is equivalent to changing the environment (here means the surface web, see Fig. \ref{fig:schematic}a) of the agent, which requires the agent to adjust to in task execution.

Stress testing is the ultimate test for model behavior and trustworthiness. In the time domain, out-of-sample (OOS) testing is typically used for temporal prediction methods \citep{hansen_equivalence_2015}. Analogous OOS assessments in the text domain include predicting future events \citep{ye_mirai_2024} or generating hypotheses \citep{zhou_hypothesis_2024} conditioned on existing (e.g. past) knowledge. We investigate the comparable problem from the tool use perspective, where the agent has access to changing internet-scale information. Because scientific breakthroughs often lead to significant knowledge updates, they are good markers for temporal knowledge progression\footnote{Although the judgement on breakthroughs are ultimately subjective and can change with time, our motivation to use them is because of their noticeable footprints on the internet.}. In this work, we aim to investigate the following research questions:
\begin{enumerate}[wide, labelindent=1em, itemsep=-2pt, topsep=1pt]
    \item[\textbf{RQ1}:] Can we manipulate agent knowledge by imposing date restrictions on the tools?
    \item[\textbf{RQ2}:] Can agents determine the optimal date-controlled version of a tool to use for a task?
\end{enumerate}

Our contributions along these directions are: (i) Formulation of tool-based stress test for time-dependent knowledge for LLM agents; (ii) Introduction of the \texttt{SciBreak} dataset containing the publication records of public-endorsed scientific breakthroughs from 2000 to 2024. (iii) Investigation of the temporal effects of LLM agent performance and behavior. Besides, we also discuss the impact of temporal information on the agent capability and usability and its implications.
% to subsequent research and applications.

\section{Related works}
\label{sec:previous}

% \paragraph{Agentic tool use in science} LLM-based agents has been used in scientific tasks such as mining information from molecular and materials science databases and publications \citep{chemcrow_2024,ghafarollahi_sciagents_2025} and collaborative problem-solving in medicine \citep{tang_medagents_2024}. The tools used in these settings are generally data processing scripts or APIs connected to external information sources \citep{wang_tools_2024}.

\paragraph{Temporal notion of LLMs} Previous works have shown that the latent space of LLMs has a direction of time \citep{gurnee_llm_2024}. Recent investigations show that model performance is affected by the lack of temporal grounding in the pre-training process \citep{zhao_stc_2024}, which can hinder the elicitation of appropriate time-sensitive knowledge at task execution. Previous works have shown LLMs often struggle with tasks that require a consistent temporal grounding \citep{qiu_tempground_2024}. The limitation can be improved with techniques such as temporally-informed chain-of-thought (CoT) prompting \citep{xiong_large_2024}.
% LLMs have been used for predicting future events \citep{ye_mirai_2024} and conducting time series analysis \citep{tang_time_2024,jin_llmtsa_2024}.

\paragraph{Out-of-sample testing} Classical and learning-based time series forecasting commonly employ temporal OOS performance tests \citep{inoue_testing_2005,hansen_equivalence_2015,cerqueira_tsa_2020} to ensure model credibility and usability. It is also relevant from an online learning perspective where data are streamed in sequentially and are subject to distribution shifts. In deep learning, OOS testing is used to provide risk-based self-certification for neural networks \citep{perez-ortiz_progress_2021,perez-ortiz_tighter_2021}. In generative models, it has been used for prompt selection \citep{perez_true_2021} and controlled generation \citep{jie_prompt-oos_2024} in language models and for quality assessment of synthetic signal generators \citep{truong_epileptic_2019}.

\section{Tool-based stress testing}
\label{sec:oott}

\begin{defn}[Date-controlled tool (DCT)]
A DCT $\mathcal{T}_t$ is a function interface $\mathcal{T}$ (base tool) with a settable parameter $t$ (upper terminal date) such that the effect of the tool at different times $t_1 \neq t_2$ are distinct, or $\mathcal{T}_{t_1} \neq \mathcal{T}_{t_2}$. The symbol $\mathcal{T}_{t_1}$ indicates that the tool was dated at $t_1$ or that it encompasses all that came before $t_1$, which is equivalent to $\mathcal{T}_{t \leqslant t_1}$. We use $\mathcal{T}_{t_1, t_2}$, or equivalently, $\mathcal{T}_{t_1 \leqslant t \leqslant t_2}$, to describe a tool assigned with a temporal window access in $t \in (t_1, t_2]$, where $t_1$ is the lower terminal date.
\end{defn}
% We adopt here the generalized definition of tools for LLM-based agents by \citet{wang_tools_2024} as a function interface.
% The output of the agent given task instruction $z$ and task-related input context $x$ is $\mathcal{M}(x;z) \circ \mathcal{T}$.
\begin{defn}[LLM agent with tools]
An LLM agent $\mathcal{A}$ with the base model $\mathcal{M}$ equipped with an invocable tool $\mathcal{T}$ is $\mathcal{A} = \mathcal{M} \circ \mathcal{T}$. A single tool invocation by the agent given input $X$ produces the trajectory $\tau_n = \{(O, R, G)_i\}_{i=1}^n$ involving the observation $O$, the reasoning trace $R$, and the action $G$. The output of the agent is described by the altered distribution
\vspace{-0.5em}
\begin{align}
    &{\Pr}_{\mathcal{A}}(Y|X) \xrightarrow[\mathcal{T}]{\,\,\textrm{single use}} {\Pr}_{\mathcal{A}}(Y|X; \tau_1),\\
    &\mathcal{T}:S \rightarrow O \,\,\Longrightarrow\,\, \mathcal{T}_t:S \rightarrow O_t. \label{eq:toolcall}
\end{align}
The tool $\mathcal{T}$ converts the source information $S$ from the environment into the observation $O$ to support agent reasoning and action. In Fig. \ref{fig:schematic}b, $S$ refers to the surface web and $O$ the ranked snippets.
\end{defn}
A web-search agent has an implicit parameter $t=t_{\max}$ (i.e. the current date) for the tool $\mathcal{T}$, but it can be modified to an arbitrary value $t < t_{\max}$, which changes the observation in Eq. \eqref{eq:toolcall}.

\begin{defn}[Tool-based stress test]
A performance test that induces stress condition by adjusting the tool parameters of an agentic system. A temporal version of the test alters the time information of tools and therefore measures the reliability of agent performance under such conditions.
\end{defn}

\section{Testing framework implementation}
\label{sec:agevals}

\paragraph{Dataset} We constructed the \texttt{SciBreak} dataset, which has a clear time-delimited footprint on the internet\textemdash scientific breakthroughs. We extended the dataset collated in \citet{wuestman_typology_2020} to the year of 2024. Each year contains up to $\sim$ 20 publications, including multiple publications contributing to one breakthrough.

\paragraph{Agent configuration} We integrated DCTs into the ReAct \citep{yao_react_2022} agent which allows interleaved thinking and action. The agents were constructed from closed-source models, including OpenAI's GPT-3.5-turbo (\texttt{gpt-3.5-turbo-0125}), GPT-4-turbo (\texttt{gpt-4-turbo-2024-04-09}) \citep{openai_gpt-4_2024}, and GPT-4o (\texttt{gpt-4o-2024-08-06}) \citep{openai_gpt-4o_2024} as the base model. For the temporal tool selection task, we also included CoT into the agent pattern (ReAct+CoT) as comparison.
% These models exhibit a variety of reasoning capabilities and are an important aspect to assess in the tasks described below.
% We used the \texttt{LangChain} family of tools for constructing the LLM agent and adapted search APIs to operate in date-controlled mode.
% and open-weight models, including \texttt{Llama3-3.1-8B} and \texttt{Deepseek-R1} in distilled formats

\paragraph{Task design and metrics} The LLM agent acted as a writing assistant and was tasked to complete the abstract of scientific papers. All evaluations were in the form of cloze tests with random masking at the word level. The agent was allowed to seek relevant information through the Google Search API with specified dates to acquire information in the form of text snippets \citep{strzelecki_direct_2020} in default ranking of the search engine. The agent then decides if the returned search results are relevant or it prefers to use its own knowledge otherwise. We modulated the information presented to the agent through the masking ratio, $\gamma$ = \#(masked words) / \#(total words), which was compared for different runs at 0.5 and 0.75, respectively.
\begin{figure*}[ht]
    \centering
    \subfloat[GPT-3.5-turbo]{
        \includegraphics[width=0.32\linewidth]{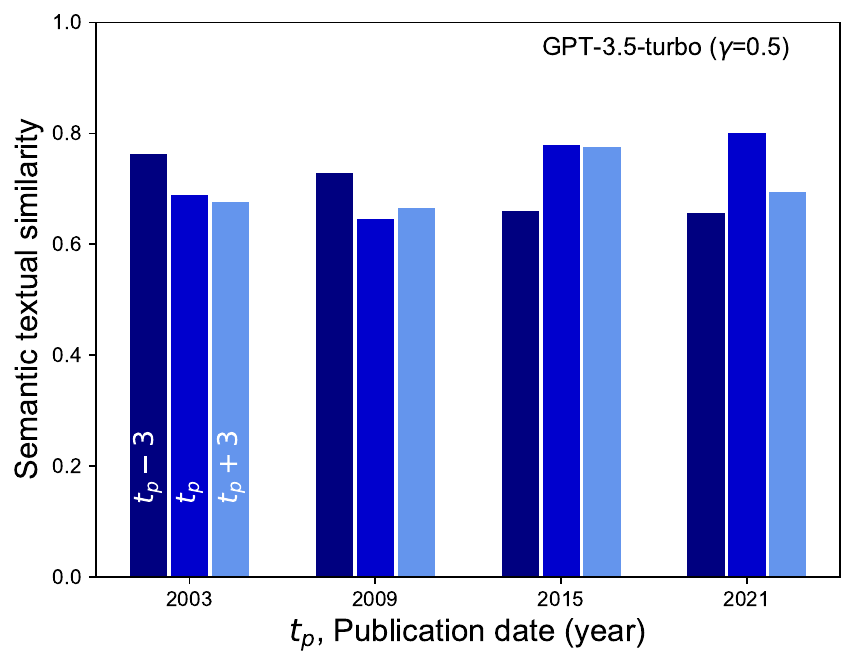}
    }
    \hfill
    \subfloat[GPT-4-turbo]{
        \includegraphics[width=0.32\linewidth]{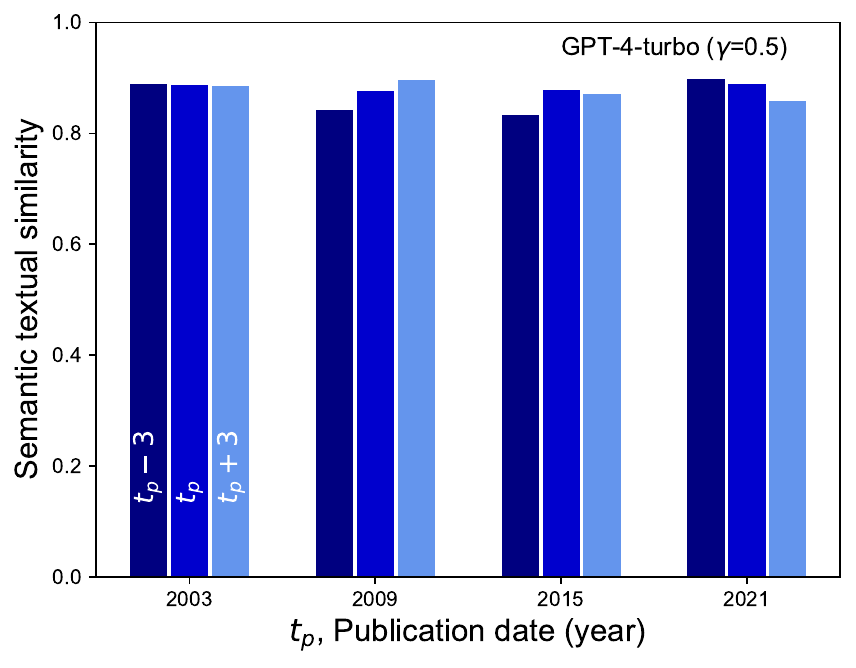}
    }
    \hfill
    \subfloat[GPT-4o]{
        \includegraphics[width=0.32\linewidth]{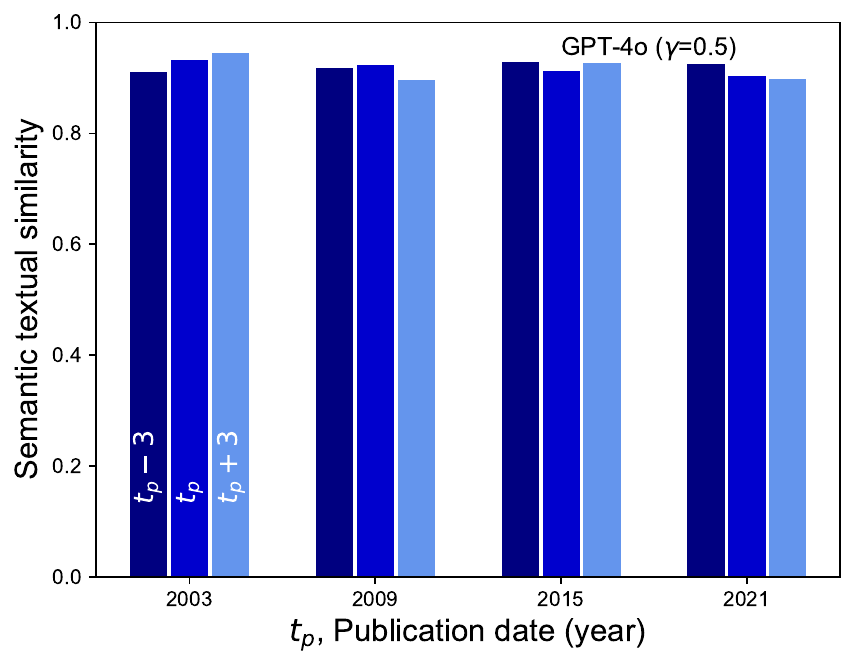}
    }
    \caption{Temporal effects of the search engine on agent performance in scientific abstract completion ($\gamma$ = 0.5).}
    \label{fig:temp_effects_g50}
\vspace{-1em}
\end{figure*}

% \vspace{-1em}
% \end{wraptable}

For \textbf{RQ1}, we evaluated how the text completion task is influenced by changing the upper terminal date of the web search ($\mathcal{T}_t$ with $t=t_p-3, t_p, t_p+3$ \text{years}) predating and postdating the time of the publication, $t_p$. For \textbf{RQ2}, we instructed the agent on temporal tool selection through CoT prompting \citep{wei_cot_2022,chu_navigate_2024}. The agent was was presented with a set $\mathcal{T}_s$ of $N$ differently date-controlled tools, $\mathcal{T}_s = \{\mathcal{T}_{t_i,t_{i}+1}\}_{i=1}^N$, each spanning the period of a year. The model need to rely on the time parameter to make decisions. For simplicity, all API web searches were done in English.

We quantify the task performance by comparing the actual version of the scientific abstract using the text overlap metric Rouge-L \citep{lin_rouge_2004} and the semantic text similarity (STS) computed with SentenceTransformer \citep{sentence-bert_2019}. The STS is the primary performance metric while Rouge-L is an indicator for verbatim completion.

% Effectively, the two tests capture both the in-sample and out-of-sample performance of the agent through by imposing DCTs.

\section{Results}

% We analyzed the agent performance and reasoning patterns, following the previous taxonomy of LLM temporal behaviors in \citet{wallat_temporal_2024}.

% We analyzed the agent performance including the patterns of agent reasoning. Our description follows the previous taxonomy on temporal behaviors of LLMs in \citet{wallat_temporal_2024}.
% A summary of the performance shift is presented in Fig. \ref{fig:bmsts_0p5}.
%  and other results are described below
% Tables \ref{tab:perf_txtcomp_g50}-\ref{tab:perf_toolsel_g75} (including two in the Appendix) show the results for the two tasks previously described. 

% \subsection{Reasoning patterns about time}
\paragraph{Reasoning about time} In our evaluation experiments, the agent's reasoning behavior related to its awareness of time (e.g. does the tool have time-appropriate utility?) is triggered in two scenarios: (i) When the web search returns nothing or little relevant information to assist task completion. The agent then proceeds to complete the task with the internal knowledge of the base LLM. (ii) In temporal tool selection, when the agent is given an explicit CoT stepwise instruction \citep{chu_navigate_2024} to direct its reasoning towards considering the relavance of information to the topic.

Table \ref{tab:perf_toolsel_g50} show that the STS increases by including CoT prompting (ReAct + CoT) than with ReAct only, when the agent by default selects $\mathcal{T}_{t_+-1, t_+}$ as the tool. Here, $t \in [t_-, t_+]$ being the date range of the tools. The agent then explores more date choices driven by its internal understanding of the scientific concepts present in the input paragraph. These behavioral characteristics allow the agent to handle non-existent and underspecified contexts in the stress test setting. The performance boost of ReAct + CoT agent pattern requires a model with sufficient reasoning capability such as GPT-4 \citep{openai_gpt-4_2024}, while for GPT-3.5, it is more prone to failure and the performance gain is reversed.
% \paragraph{Temporal effects on agent knowledge} Agents can \textbf{handle non-existent knowledge}. This is crucial. Besides, we also see agent's capability of \textbf{handling underspecified knowledge}. More capable models are able to provide more structured reasoning and piece together a fuller picture given limited information of the abstract.

% \captionsetup[subfloat]{captionskip=3pt}

% \begin{figure}[htbp]
%     \centering
%     \includegraphics[width=\linewidth]{figs/Test_outcome_GPT3p5.pdf}
%     \caption{Agent performance (assessed by STS) on the \texttt{SciBreak} dataset using date-controlled search tools at a masking ratio of 0.5.}
%     \label{fig:bmsts_0p5}
% \end{figure}
\begin{table}[htbp]
% \begin{wraptable}{R}{0.55\textwidth}
% \vspace{-3.5em}
\small
% \scriptsize
\setstretch{1.1}
\caption{Performance of LLM agents on text completion ($\gamma = 0.5$) with temporal tool selection.}
\vspace{-1em}
% \label{sample-table}
\begin{center}
\begin{tabular}{@{\hskip-2pt}c@{\hskip0pt}|@{\hskip3pt}c@{\hskip4pt}c@{\hskip5pt}c@{\hskip8pt}c@{\hskip5pt}c@{\hskip0pt}}
\toprule
\multirow{2}{*}{\begin{tabular}{c}\textbf{Agent}\\\textbf{model}\end{tabular}} & \multirow{2}{*}{\begin{tabular}{c}\textbf{Agent}\\\textbf{pattern}\end{tabular}} & \multicolumn{2}{c}{\textbf{Pub. in 2003}} & \multicolumn{2}{c}{\textbf{Pub. in 2015}} \\
\cmidrule(lr){3-4} \cmidrule(lr){5-6} 
& & \textbf{Rouge-L} & \textbf{STS} & \textbf{Rouge-L} & \textbf{STS} \\
\midrule
\multirow{2}{*}{\begin{tabular}{c}GPT-3.5\\-turbo\end{tabular}}
% $t_p+2$ yrs     &  & \\
& ReAct       & 0.447 & \textbf{0.732} & 0.447 & \textbf{0.794} \\
% $t_p-2$ yrs     &  & \\
& ReAct + CoT     & 0.487 & 0.640 & 0.518 & 0.607 \\
% $t_p-10$ yrs    &  & \\
\midrule
\multirow{2}{*}{\begin{tabular}{c}GPT-4\\-turbo\end{tabular}} 
& ReAct & 0.635 & 0.888 & 0.588 & 0.783 \\
& ReAct + CoT & 0.649 & \textbf{0.897} & 0.644 & \textbf{0.859} \\
% \midrule
% \multirow{3}{*}{GPT-4o} & $t_p-3$ & 0.735 & 0.911 & 0.727 & \textbf{0.928} \\
% & $t_p$       & 0.741 & 0.931 & 0.744 & 0.912 \\
% & $t_p+3$     & 0.733 & \textbf{0.944} & 0.701 & \textbf{0.928} \\
\bottomrule
\end{tabular}
\label{tab:perf_toolsel_g50}
\end{center}
\vspace{-1em}
\end{table}

\paragraph{Temporal effects across models and masking} High-capacity models with pronounced reasoning capabilities are capable of examining tool dates. The task evaluated the model capability with two different levels of text masking. In Figs. \ref{fig:temp_effects_g50}-\ref{fig:temp_effects_g75} and Tables \ref{tab:perf_txtcomp_g50}-\ref{tab:perf_toolsel_g75}, the outcome contains two major trends: (i) The more advanced models can recover more of the missing semantic content in the masked input, as indicated by the significant increase of STS from LLM agents based on GPT-3.5 to GPT-4o \citep{openai_gpt-4o_2024}. (ii) There is noticeable variability of agent performance between knowledge generated more recently than before 2010. (iii) For the same model, varying the masking ratio of input largely preserves the date sensitivity in the performance. Similar time-dependent performance change has been described in a different context for LLMs \citep{zhao_stc_2024}. Overall, the temporal effects are less severe in more capable models.
% than their lesser counterparts.

\section{Discussion}
% search is subject to similar supplying date-restricted tools is a potential way to control agent capabilities in high-capacity models.
% A comparison between the reasoning traces of a higher- and lower-performing agents show different sensitivity to externalities.

\paragraph{Agent vulnerability and tool-based control} Our work shows that agents with access to external tools are subject to manipulation by corrupted tools \citep{ye_toolsword_2024} to compromise their generated information for knowledge-intensive domains, extending previous example on misinformation in LLMs \citep{han_medicalmisinfo_2024}. We provide evidence that agentic reasoning and model capabilities can counter the limited information quality of search engines. In agentic search, carefully designed controls will allow filtering of unreliable information and improve agent performance. Imposing date restriction on search is similar to reranking and partial deletion of the search results. Therefore, agent designs with verification of content freshness and temporality will ensure more reliable use.

\paragraph{Robustness and reproducibility of agentic systems} Agentic systems for scientific problems should adapt to different levels of prior knowledge available to the domain \citep{vinuesa_opportunities_2024}. From the robustness viewpoint, temporal shifts can be counteracted through the use of external resources. Task-oriented requirements specification \citep{xian_robustness_2025} is useful for improving the usability and avoid unnecessary artifacts from model imperfections and the reliability of external tooling and information sources. From the reproducibility viewpoint, agentic tool use should always incorporate essential information of the key parameters. Our work indicates that more research is needed in principled maintenance of agentic frameworks under constant updates of external resources to facilitate reliable agent design \citep{kapoor_agent_matters_2024}.

\section*{Limitations}

Our work is focused on models with tool-calling and reasoning capabilities, yet the phenomenon demonstrated here has equivalents in less capable models not investigated here. The test examples we chose simulates a realistic application setting of an agentic writing assistant, yet such effect could already manifest in more ordinary tasks such as knowledge-related question answering or in malicious settings where bad actors are trying to pollute the information system (e.g. internet or proprietary databases) through more elaborate search engine manipulation. We also didn't investigate the scenario where the LLM agent has possession of a proprietary tool (e.g. for fact-checking) independent of web search, which could be an alternative way to improve performance.

\section*{Ethics statement}

The present work illustrates the importance of temporal factors when working with LLM agents that have access to the internet. Our results provide an initial assessment of the factors that can influence an agentic writing assistant's ability to properly utilize time-bounded search results in its reasoning process. We acknowledge that reliance on time-bounded search results presents ethical considerations related to misinformation, data freshness, and accuracy. Agents may misinterpret outdated or contextually misaligned information, leading to erroneous conclusions. Furthermore, temporal biases in search tools, such as the prioritization of newer content over historically relevant sources, can skew results, potentially reinforcing recency bias or omitting crucial context.

% \section*{Acknowledgments}

% R.P.X. thanks N. Rethmeier for helpful discussions. R.A.-A. would like to acknowledge funding from Weill Neurohub.

% Bibliography entries for the entire Anthology, followed by custom entries
% \bibliography{acl_latex}
% Custom bibliography entries only

\appendix
% \label{sec:appendix}
\clearpage
\section{Agentic task}
The web-search agent is configured according to the ReAct architecture using a helpful assistant system prompt (``You are a helpful writing assistant.''). The instruction prompt is as follows. 

\begin{tcolorbox}[colback=lightgray!20, boxrule=1pt, left=1.5mm, right=1.5mm, top=1.5mm, bottom=1.5mm, rounded corners, coltitle=Black, drop shadow=black!50!white, fonttitle=\bfseries]
\setstretch{1.1}
\small{
Instruction: Browse the internet using keywords or phrases in the following paragraph with masked text. Make use of the search results to fill in each [UNK] with a word or punctuation. Output your final results after Final Answer: to indicate the beginning of your completed text. Use your own judgement to decide what information from the search results are useful. If nothing is useful, then try to complete the task with your own knowledge.

Requirements: Use the given parameters in the tools to solve the following problem and don't reset them. Don't change the number of arguments supplied to the tool you use. \\ Masked text:{} ... [\texttt{the masked text}] ...
}
\end{tcolorbox}
In our experiments, \texttt{the masked text} is replaced with the masked scientific abstracts. The instruction prompt contains task description and to suppress undesired agent behaviors that can cause error in execution. In our empirical investigation, we also found that the reasoning process of tool-calling agents has a tendency to reset the date parameter. For the experiments, we added an instruction to specifically forbid that behavior.
% while the space after ``Completed text'' is left blank for the agent to complete with a modified version

\section{Dataset preprocessing}
The \texttt{SciBreak} dataset is partly based on peer-reviewed publications collated and categorized in \citet{wuestman_typology_2020}, including the annual top-ten-ranked scientific breakthroughs from mid-1990s till 2012 collated by the journal Science at the end of each year. The publications are drawn from various journals in the physical, biomedical, and engineering sciences, which constitute the scope of the ranking. We chose records from year 2000 to 2012 and extended to year 2024 by self-curating the extra years of ranked publications from the published tally in each year. The links to the yearly breakdown is provided as follows: \href{https://www.science.org/content/article/sciences-top-10-breakthroughs-2013}{2013}, \href{https://www.science.org/content/article/breakthrough-year-top-10-scientific-achievements-2014}{2014}, \href{https://www.science.org/content/article/and-science-s-2015-breakthrough-year}{2015} (12), \href{https://vis.sciencemag.org/breakthrough2018/finalists/}{2018} (17), \href{https://www.science.org/content/article/breakthrough-2021}{2021} (14), \href{https://www.science.org/content/article/breakthrough-2024}{2024} (11).

We collected the abstracts of the associated publications through web scraping from the public databases PubMed\footnote{\url{https://pubmed.ncbi.nlm.nih.gov/}} and SAO/NASA ADS Abstract Service\footnote{\url{https://ui.adsabs.harvard.edu/}} using the Digital Object Identifiers of the publications, which are also provided in the dataset.

% The paper abstracts are accessible on the internet when the tool with appropriate date is used, otherwise the search returns nothing or outdated or irrelevant information. If the tool with the wrong date is selected, then the agent can still complete the task by inferring from past events or potentially from the pre-training data of the base model of the agent. 

% \section{Masking effect on text data}
% As a sanity check, we investigated the relationship between the STS and the masking raio for the scientific abstract data to assess the retention of the semantic content before and after masking. Fig. \ref{fig:sem_retent} shows the effect using an example.
% \begin{figure}[htbp]
%     \centering
%     \includegraphics[width=\linewidth]{figs/semantic_retention.pdf}
%     \caption{Relation between masking ratio and STS. Error bars are the standard deviation.}
%     \label{fig:sem_retent}
% \end{figure}

\section{Extended results}
\begin{table}[htbp]
% \begin{wraptable}{R}{0.55\textwidth}
\vspace{-5pt}
\small
% \scriptsize
\setstretch{1.1}
\caption{Performance of ReAct-style LLM agents on text completion ($\gamma = 0.5$, see Fig. \ref{fig:temp_effects_g50}) following web search with DCTs. For the row of Input, the metrics are computed between the input and the ground truth.}
% \label{sample-table}
\vspace{-1em}
\begin{center}
\begin{tabular}{@{\hskip-2pt}c@{\hskip0pt}|@{\hskip0pt}c@{\hskip5pt}c@{\hskip5pt}c@{\hskip8pt}c@{\hskip5pt}c@{\hskip0pt}}
\toprule
\multirow{2}{*}{\begin{tabular}{c}\textbf{Agent}\\\textbf{model}\end{tabular}} & \multirow{2}{*}{\begin{tabular}{c}\textbf{$\mathcal{T}_t$ cut-off}\\\textbf{($t$ years)}\end{tabular}} & \multicolumn{2}{c}{\textbf{Pub. in 2003}} & \multicolumn{2}{c}{\textbf{Pub. in 2015}} \\
\cmidrule(lr){3-4} \cmidrule(lr){5-6} 
& & \textbf{Rouge-L} & \textbf{STS} & \textbf{Rouge-L} & \textbf{STS} \\
\midrule
Input & \textemdash & 0.486 & 0.657 & 0.488 & 0.658 \\
\midrule
\multirow{3}{*}{\begin{tabular}{c}GPT-3.5\\-turbo\end{tabular}} & $t_p-3$ & 0.481 & \textbf{0.764} & 0.623 & 0.659 \\
% $t_p+2$ yrs     &  & \\
& $t_p$       & 0.462 & 0.689 & 0.438 & \textbf{0.779} \\
% $t_p-2$ yrs     &  & \\
& $t_p+3$     & 0.427 & 0.677 & 0.546 & 0.774 \\
% $t_p-10$ yrs    &  & \\
\midrule
\multirow{3}{*}{\begin{tabular}{c}GPT-4\\-turbo\end{tabular}} & $t_p-3$ & 0.627 & \textbf{0.889} & 0.601 & 0.833 \\
& $t_p$       & 0.641 & 0.887 & 0.629 & \textbf{0.879} \\
& $t_p+3$     & 0.614 & 0.886 & 0.624 & 0.871 \\
\midrule
\multirow{3}{*}{GPT-4o} & $t_p-3$ & 0.735 & 0.911 & 0.727 & \textbf{0.928} \\
& $t_p$       & 0.741 & 0.931 & 0.744 & 0.912 \\
& $t_p+3$     & 0.733 & \textbf{0.944} & 0.701 & \textbf{0.928} \\
\bottomrule
\end{tabular}
\label{tab:perf_txtcomp_g50}
\end{center}
\vspace{-1em}
\end{table}

\begin{table}[hbt!]
% \begin{wraptable}{R}{0.55\textwidth}
\vspace{-0.5em}
\small
% \scriptsize
\setstretch{1.1}
\caption{Performance of ReAct-style LLM agents on text completion ($\gamma = 0.75$, see Fig. \ref{fig:temp_effects_g75}) following web search with DCTs. For the row of Input, the metrics are computed between the input and the ground truth.}
% \label{sample-table}
\begin{center}
\begin{tabular}{@{\hskip-2pt}c@{\hskip0pt}|@{\hskip0pt}c@{\hskip5pt}c@{\hskip5pt}c@{\hskip8pt}c@{\hskip5pt}c@{\hskip0pt}}
\toprule
\multirow{2}{*}{\begin{tabular}{c}\textbf{Agent}\\\textbf{model}\end{tabular}} & \multirow{2}{*}{\begin{tabular}{c}\textbf{$\mathcal{T}_t$ cut-off}\\\textbf{($t$ years)}\end{tabular}} & \multicolumn{2}{c}{\textbf{Pub. in 2003}} & \multicolumn{2}{c}{\textbf{Pub. in 2015}} \\
\cmidrule(lr){3-4} \cmidrule(lr){5-6} 
& & \textbf{Rouge-L} & \textbf{STS} & \textbf{Rouge-L} & \textbf{STS} \\
\midrule
Input & \textemdash & 0.486 & 0.657 & 0.488 & 0.658 \\
\midrule
\multirow{3}{*}{\begin{tabular}{c}GPT-3.5\\-turbo\end{tabular}} & $t_p-3$ & 0.418 & \textbf{0.636} & 0.286 & 0.525 \\
% $t_p+2$ yrs     &  & \\
& $t_p$       & 0.279 & 0.590 & 0.274 & \textbf{0.626} \\
% $t_p-2$ yrs     &  & \\
& $t_p+3$     & 0.285 & 0.602 & 0.309 & 0.586 \\
% $t_p-10$ yrs    &  & \\
\midrule
\multirow{3}{*}{\begin{tabular}{c}GPT-4\\-turbo\end{tabular}} & $t_p-3$ & 0.324 & 0.729 & 0.325 & \textbf{0.767} \\
& $t_p$       & 0.341 & 0.706 & 0.376 & 0.762 \\
& $t_p+3$     & 0.390 & \textbf{0.826} & 0.319 & 0.705 \\
\midrule
\multirow{3}{*}{GPT-4o} & $t_p-3$ & 0.478 & \textbf{0.867} & 0.405 & 0.796 \\
& $t_p$       & 0.440 & 0.842 & 0.448 & 0.842 \\
& $t_p+3$     & 0.446 & 0.846 & 0.458 & \textbf{0.863} \\
\bottomrule
\end{tabular}
\label{tab:perf_txtcomp_g75}
\end{center}
\vspace{-1em}
\end{table}
\begin{table}[H]
% \begin{wraptable}{R}{0.55\textwidth}
% \vspace{-3.5em}
\small
% \scriptsize
\setstretch{1.1}
\caption{Performance of LLM agents on text completion ($\gamma = 0.75$) with temporal tool selection.}
% \label{sample-table}
\begin{center}
\begin{tabular}{@{\hskip-2pt}c@{\hskip0pt}|@{\hskip3pt}c@{\hskip4pt}c@{\hskip5pt}c@{\hskip8pt}c@{\hskip5pt}c@{\hskip0pt}}
\toprule
\multirow{2}{*}{\begin{tabular}{c}\textbf{Agent}\\\textbf{model}\end{tabular}} & \multirow{2}{*}{\begin{tabular}{c}\textbf{Agent}\\\textbf{pattern}\end{tabular}} & \multicolumn{2}{c}{\textbf{Pub. in 2003}} & \multicolumn{2}{c}{\textbf{Pub. in 2015}} \\
\cmidrule(lr){3-4} \cmidrule(lr){5-6} 
& & \textbf{Rouge-L} & \textbf{STS} & \textbf{Rouge-L} & \textbf{STS} \\
\midrule
\multirow{2}{*}{\begin{tabular}{c}GPT-3.5\\-turbo\end{tabular}} & ReAct       & 0.297 & \textbf{0.604} & 0.438 & \textbf{0.779} \\
% $t_p-2$ yrs     &  & \\
& ReAct + CoT     & 0.212 & 0.416 & 0.546 & 0.774 \\
% $t_p-10$ yrs    &  & \\
\midrule
\multirow{2}{*}{\begin{tabular}{c}GPT-4\\-turbo\end{tabular}}
& ReAct       & 0.343 & 0.756 & 0.311 & 0.666 \\
& ReAct + CoT     & 0.447 & \textbf{0.796} & 0.304 & \textbf{0.693} \\
% & ReAct + $\mathcal{T}_{t_{\text{rand}}}$ & 0.481 & \textbf{0.764} & 0.525 & 0.788 \\

% & ReAct + $\mathcal{T}_{t_{\text{rand}}}$ & 0.627 &  &  &  \\
% \midrule
% \multirow{3}{*}{GPT-4o} & $t_p-3$ & 0.735 & 0.911 & 0.727 & \textbf{0.928} \\
% & $t_p$       & 0.741 & 0.931 & 0.744 & 0.912 \\
% & $t_p+3$     & 0.733 & \textbf{0.944} & 0.701 & \textbf{0.928} \\
\bottomrule
\end{tabular}
\label{tab:perf_toolsel_g75}
\end{center}
\vspace{-1em}
\end{table}
\begin{figure*}[ht]
    \centering
    \subfloat[GPT-3.5-turbo]{
        \includegraphics[width=0.32\linewidth]{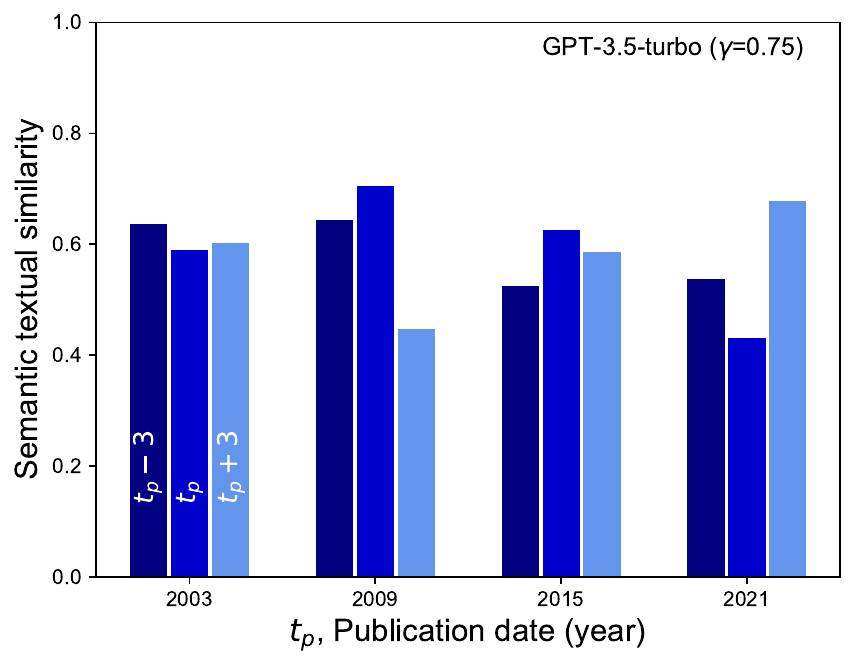}
    }
    \hfill
    \subfloat[GPT-4-turbo]{
        \includegraphics[width=0.32\linewidth]{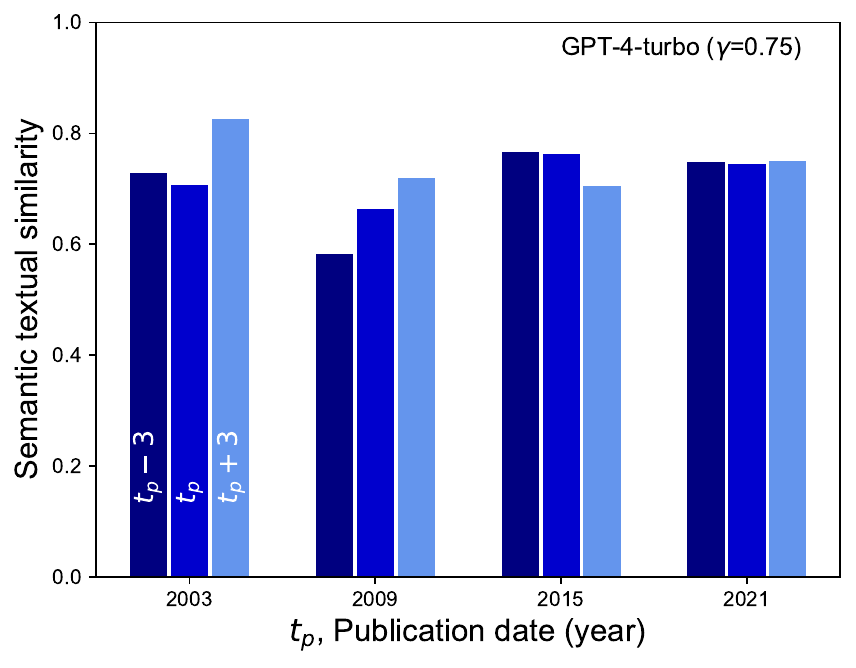}
    }
    \hfill
    \subfloat[GPT-4o]{
        \includegraphics[width=0.32\linewidth]{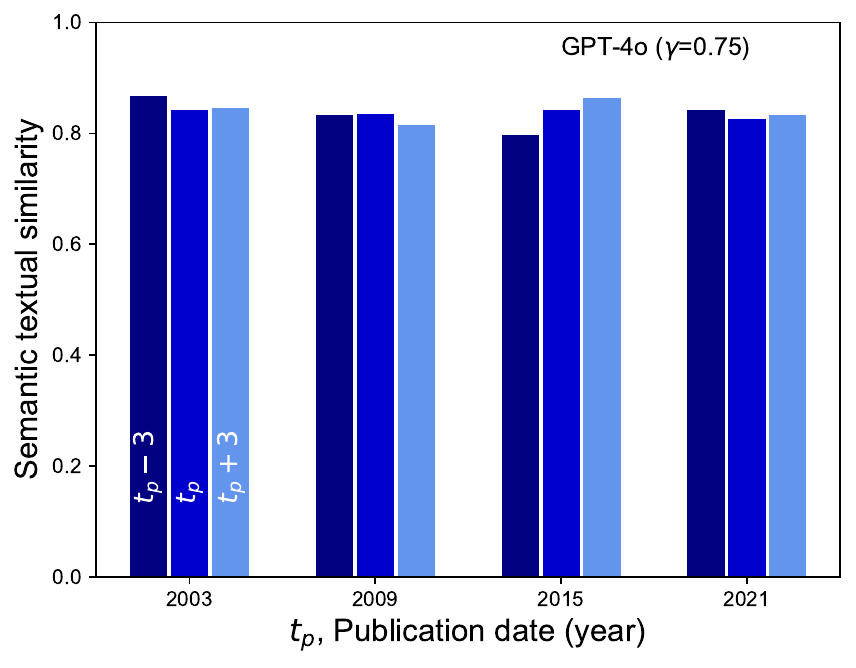}
    }
    \caption{Temporal effects of the search engine on agent performance in scientific abstract completion ($\gamma$ = 0.75).}
    \label{fig:temp_effects_g75}
\vspace{-1em}
\end{figure*}
\noindent Extended results for \textbf{RQ1} include Table \ref{tab:perf_txtcomp_g50}, which shows a portion of the results in Fig. \ref{fig:temp_effects_g50}, and Table \ref{tab:perf_txtcomp_g75}, which is similarly related to Fig. \ref{fig:temp_effects_g75}. Results in Table \ref{tab:perf_toolsel_g75} are the extension for temporal tool selection (\textbf{RQ2}) with a different masking ratio (compare Table \ref{tab:perf_toolsel_g50}). These are from experiments carried out at a masking ratio of $\gamma = 0.75$. Although most of the performance metrics trend lower than the conditions in Tables \ref{tab:perf_txtcomp_g50}-\ref{tab:perf_toolsel_g50}, when $\gamma = 0.5$, similar characteristics hold for the time dependence of the STS for publications appeared in more recently (since 2010s) and those before. In the temporal tool selection task, the ReAct + CoT pattern provides performance gain over ReAct-only agents.
\begin{figure}[hbtp!]
% \vspace{-1em}
\subfloat[No date restriction on tool.]{
\begin{minipage}{\linewidth}
\begin{tcolorbox}[colback=TealBlue!20, boxrule=1pt, left=2mm, right=2mm, top=2mm, bottom=2mm, rounded corners, coltitle=Black, drop shadow=black!50!white, fonttitle=\bfseries]
\setstretch{1.1}
\small{
\begin{enumerate}[wide, labelindent=0pt]
\item[\textbf{Observation}:] \textsf{None}
\item[\textbf{Thought}:] \textsf{The search results provide detailed information about the discovery and genetic analysis of Denisovans. \ul{I can use this information to fill in the blanks} in the masked text. The text seems to describe the discovery of Denisovan remains in Denisova Cave, Siberia, and their genetic relationship to Neanderthals and modern humans.}
\end{enumerate}
}
\end{tcolorbox}
\vspace{1pt}
\end{minipage}}
\hfill
\subfloat[Tool restricted to date before event.]{
\begin{minipage}{\linewidth}
\begin{tcolorbox}[colback=Tan!20, boxrule=1pt, left=2mm, right=2mm, top=2mm, bottom=2mm, rounded corners, coltitle=Black, drop shadow=black!50!white, fonttitle=\bfseries]
\setstretch{1.1}
\small{
\begin{enumerate}[wide, labelindent=0pt]
\item[\textbf{Observation}:] \textsf{The search did not return any results. This could be because the discovery of Denisovans and the relevant genetic analysis occurred after 2004. \ul{Therefore, I need to adjust my approach to fill in the blanks based on general knowledge} available up to 2004 about ancient hominins, their discoveries, and genetic analysis techniques.}
\item[\textbf{Thought}:] \textsf{\underline{Since the Denisovan discovery itself was} \underline{not known before 2004, I will use general knowledge} about ancient DNA studies and hominin discoveries in Siberia to fill in the blanks.}
\end{enumerate}
}
\end{tcolorbox}
\vspace{1pt}
\end{minipage}}
\caption{Example reasoning paths (emphasized by underlines) from the LLM agent before and after imposing a date restriction on the tool. The example here uses the discovery of Denisovan hominins. Important parts of the verbalized reasoning are underlined.}
\label{fig:denisovan}
\vspace{-0.5em}
\end{figure}

\section{Examples of temporal awareness}
The example in Fig. \ref{fig:denisovan} includes the typical reasoning trace of the LLM agent put under testing to the breakthrough discovery of Denisovan hominins (an ancestor of modern humans) around 2008, which became widely reported in the English media a couple of years later thanks to the major scientific publication \citep{reich_genetic_2010} and contributed significantly to Svante Pääbo's Nobel Prize in 2022.

The Denisova cave in Siberia exists as a geographical name for much longer on the internet, but primarily in the Russian language, so largely inaccessible through English language search before 2008. Moreover, Denisova is used as a surname, which appears upon search in English. However, neither of these facts inform the model about potential content in the masked text about the scientific discovery that was consolidated by genomic sequencing. When the clock of the search engine is set to before 2008, the LLM agent attempted to confront the absence of results and reasoned that the work was not known before the cut-off date of the search and instead switched to using its parametric knowledge to complete the text. If the search clock was unset, then the information is readily available, as compared in Fig. \ref{fig:denisovan}.

% \section{Thought processes of agents in temporal reasoning}
% \paragraph{Temporal awareness} Agents reflect on the unavailability of information and found that it is the results of the date parameter applied to the search API.

% \paragraph{Tool version selection} We provide here some thought processes the model generated during inference time.
% \begin{tcolorbox}[colback=lightgray!20, boxrule=0pt, left=2mm, right=2mm, top=2mm, bottom=2mm, sharp corners, coltitle=Black, drop shadow=black!50!white, fonttitle=\bfseries]
% \setstretch{0.9}
% \small{
% The chatbot which can generate a medication request for patients has been extensively tested for robustness under the following scenarios:
% \begin{enumerate}[wide, labelindent=0pt]
% \item Up to 12 turns of conversation on topics related to primary care medication
% \item Knowledge of specialized vocabulary, including products of major pharmaceutical vendors
% \item Partly missing patient information (may refuse the task if patient is unwilling to provide sufficient information)
% \item Drug combinations, recommended intake quantities, and dosage limits
% \item Adverse drug interactions and patient history
% \item Stock availability and provider preference
% \item Limited prescription authority (will refuse the task if asked for non-OTC or prescription drugs)
% \item Paraphrasing and typo in instruction prompt
% \item Off-topic requests and non-existent drugs
% \end{enumerate}
% }
% \end{tcolorbox}

\bibliography{acl_latex}

\begin{thebibliography}{50}
\providecommand{\natexlab}[1]{#1}

\bibitem[{Aral and Dhillon(2021)}]{aral_digital_2021}
Sinan Aral and Paramveer~S. Dhillon. 2021.
\newblock \href {https://doi.org/10.1287/mnsc.2020.3650} {Digital {Paywall} {Design}: {Implications} for {Content} {Demand} and {Subscriptions}}.
\newblock \emph{Management Science}, 67(4):2381--2402.
\newblock Publisher: INFORMS.

\bibitem[{Arbesman(2013)}]{arbesman_half-life_2013}
Samuel Arbesman. 2013.
\newblock \emph{The {Half}-{Life} of {Facts}: {Why} {Everything} {We} {Know} {Has} an {Expiration} {Date}}, reprint edition.
\newblock Penguin Publishing Group.

\bibitem[{Bousetouane(2025)}]{bousetouane_agentic_2025}
Fouad Bousetouane. 2025.
\newblock \href {https://doi.org/10.48550/arXiv.2501.00881} {Agentic {Systems}: {A} {Guide} to {Transforming} {Industries} with {Vertical} {AI} {Agents}}.
\newblock \emph{arXiv preprint}.
\newblock ArXiv:2501.00881 [cs].

\bibitem[{Cerqueira et~al.(2020)Cerqueira, Torgo, and Mozetič}]{cerqueira_tsa_2020}
Vitor Cerqueira, Luis Torgo, and Igor Mozetič. 2020.
\newblock \href {https://doi.org/10.1007/s10994-020-05910-7} {Evaluating time series forecasting models: an empirical study on performance estimation methods}.
\newblock \emph{Machine Learning}, 109(11):1997--2028.

\bibitem[{Chu et~al.(2024)Chu, Chen, Chen, Yu, He, Wang, Peng, Liu, Qin, and Liu}]{chu_navigate_2024}
Zheng Chu, Jingchang Chen, Qianglong Chen, Weijiang Yu, Tao He, Haotian Wang, Weihua Peng, Ming Liu, Bing Qin, and Ting Liu. 2024.
\newblock \href {https://doi.org/10.18653/v1/2024.acl-long.65} {Navigate through {Enigmatic} {Labyrinth} {A} {Survey} of {Chain} of {Thought} {Reasoning}: {Advances}, {Frontiers} and {Future}}.
\newblock In \emph{Proceedings of the 62nd {Annual} {Meeting} of the {Association} for {Computational} {Linguistics} ({Volume} 1: {Long} {Papers})}, pages 1173--1203, Bangkok, Thailand. Association for Computational Linguistics.

\bibitem[{Epstein and Robertson(2015)}]{epstein_search_2015}
Robert Epstein and Ronald~E. Robertson. 2015.
\newblock \href {https://doi.org/10.1073/pnas.1419828112} {The search engine manipulation effect ({SEME}) and its possible impact on the outcomes of elections}.
\newblock \emph{Proceedings of the National Academy of Sciences}, 112(33):E4512--E4521.
\newblock Publisher: National Academy of Sciences.

\bibitem[{Gao et~al.(2024)Gao, Fang, Huang, Giunchiglia, Noori, Schwarz, Ektefaie, Kondic, and Zitnik}]{gao_empowering_2024}
Shanghua Gao, Ada Fang, Yepeng Huang, Valentina Giunchiglia, Ayush Noori, Jonathan~Richard Schwarz, Yasha Ektefaie, Jovana Kondic, and Marinka Zitnik. 2024.
\newblock \href {https://doi.org/10.1016/j.cell.2024.09.022} {Empowering biomedical discovery with {AI} agents}.
\newblock \emph{Cell}, 187(22):6125--6151.

\bibitem[{Gurnee and Tegmark(2024)}]{gurnee_llm_2024}
Wes Gurnee and Max Tegmark. 2024.
\newblock \href {https://openreview.net/forum?id=jE8xbmvFin} {Language {Models} {Represent} {Space} and {Time}}.
\newblock In \emph{The Twelfth International Conference on Learning Representations}.

\bibitem[{Han et~al.(2024)Han, Nebelung, Khader, Wang, Müller-Franzes, Kuhl, Försch, Kleesiek, Haarburger, Bressem, Kather, and Truhn}]{han_medicalmisinfo_2024}
Tianyu Han, Sven Nebelung, Firas Khader, Tianci Wang, Gustav Müller-Franzes, Christiane Kuhl, Sebastian Försch, Jens Kleesiek, Christoph Haarburger, Keno~K. Bressem, Jakob~Nikolas Kather, and Daniel Truhn. 2024.
\newblock \href {https://doi.org/10.1038/s41746-024-01282-7} {Medical large language models are susceptible to targeted misinformation attacks}.
\newblock \emph{npj Digital Medicine}, 7(1):1--9.
\newblock Publisher: Nature Publishing Group.

\bibitem[{Hansen and Timmermann(2015)}]{hansen_equivalence_2015}
Peter~Reinhard Hansen and Allan Timmermann. 2015.
\newblock \href {https://doi.org/10.3982/ECTA10581} {Equivalence {Between} {Out}-of-{Sample} {Forecast} {Comparisons} and {Wald} {Statistics}}.
\newblock \emph{Econometrica}, 83(6):2485--2505.

\bibitem[{Højmark et~al.(2024)Højmark, Pimpale, Panickssery, Hobbhahn, and Scheurer}]{hojmark_analyzing_2024}
Axel Højmark, Govind Pimpale, Arjun Panickssery, Marius Hobbhahn, and Jérémy Scheurer. 2024.
\newblock \href {https://openreview.net/forum?id=mwTemAaHTk&referrer=%5Bthe%20profile%20of%20Arjun%20Panickssery%5D(%2Fprofile%3Fid%3D~Arjun_Panickssery1)} {Analyzing {Probabilistic} {Methods} for {Evaluating} {Agent} {Capabilities}}.
\newblock In \emph{Workshop on Socially Responsible Language Modelling Research}.

\bibitem[{Inoue and Kilian(2005)}]{inoue_testing_2005}
Atsushi Inoue and Lutz Kilian. 2005.
\newblock \href {https://doi.org/10.1081/ETC-200040785} {In-{Sample} or {Out}-of-{Sample} {Tests} of {Predictability}: {Which} {One} {Should} {We} {Use}?}
\newblock \emph{Econometric Reviews}, 23(4):371--402.
\newblock Publisher: Taylor \& Francis.

\bibitem[{Jie et~al.(2024)Jie, Meng, Shang, Jiang, and Liu}]{jie_prompt-oos_2024}
Renlong Jie, Xiaojun Meng, Lifeng Shang, Xin Jiang, and Qun Liu. 2024.
\newblock \href {https://doi.org/10.18653/v1/2024.findings-acl.63} {Prompt-{Based} {Length} {Controlled} {Generation} with {Multiple} {Control} {Types}}.
\newblock In \emph{Findings of the {Association} for {Computational} {Linguistics}: {ACL} 2024}, pages 1067--1085, Bangkok, Thailand. Association for Computational Linguistics.

\bibitem[{Kapoor et~al.(2024)Kapoor, Stroebl, Siegel, Nadgir, and Narayanan}]{kapoor_agent_matters_2024}
Sayash Kapoor, Benedikt Stroebl, Zachary~S. Siegel, Nitya Nadgir, and Arvind Narayanan. 2024.
\newblock \href {https://doi.org/10.48550/arXiv.2407.01502} {{AI} {Agents} {That} {Matter}}.
\newblock \emph{arXiv preprint}.
\newblock ArXiv:2407.01502 [cs].

\bibitem[{Lau and Coiera(2007)}]{lau_people_2007}
Annie~Y.S. Lau and Enrico~W. Coiera. 2007.
\newblock \href {https://doi.org/10.1197/jamia.M2411} {Do {People} {Experience} {Cognitive} {Biases} while {Searching} for {Information}?}
\newblock \emph{Journal of the American Medical Informatics Association}, 14(5):599--608.

\bibitem[{Lawrence and Giles(1998)}]{lawrence_searching_1998}
Steve Lawrence and C.~Lee Giles. 1998.
\newblock \href {https://doi.org/10.1126/science.280.5360.98} {Searching the {World} {Wide} {Web}}.
\newblock \emph{Science}, 280(5360):98--100.
\newblock Publisher: American Association for the Advancement of Science.

\bibitem[{Lazaridou et~al.(2021)Lazaridou, Kuncoro, Gribovskaya, Agrawal, Liska, Terzi, Gimenez, d'Autume, Kočiský, Ruder, Yogatama, Cao, Young, and Blunsom}]{lazaridou_mind_2021}
Angeliki Lazaridou, Adhiguna Kuncoro, Elena Gribovskaya, Devang Agrawal, Adam Liska, Tayfun Terzi, Mai Gimenez, Cyprien de~Masson d'Autume, Tomáš Kočiský, Sebastian Ruder, Dani Yogatama, Kris Cao, Susannah Young, and Phil Blunsom. 2021.
\newblock \href {https://openreview.net/forum?id=73OmmrCfSyy} {Mind the {Gap}: {Assessing} {Temporal} {Generalization} in {Neural} {Language} {Models}}.
\newblock In \emph{Advances in Neural Information Processing Systems}.

\bibitem[{Lin(2004)}]{lin_rouge_2004}
Chin-Yew Lin. 2004.
\newblock \href {https://aclanthology.org/W04-1013/} {{ROUGE}: {A} {Package} for {Automatic} {Evaluation} of {Summaries}}.
\newblock In \emph{Text {Summarization} {Branches} {Out}}, pages 74--81, Barcelona, Spain. Association for Computational Linguistics.

\bibitem[{Mialon et~al.(2023)Mialon, Dessi, Lomeli, Nalmpantis, Pasunuru, Raileanu, Roziere, Schick, Dwivedi-Yu, Celikyilmaz, Grave, LeCun, and Scialom}]{mialon_augmented_2023}
Grégoire Mialon, Roberto Dessi, Maria Lomeli, Christoforos Nalmpantis, Ramakanth Pasunuru, Roberta Raileanu, Baptiste Roziere, Timo Schick, Jane Dwivedi-Yu, Asli Celikyilmaz, Edouard Grave, Yann LeCun, and Thomas Scialom. 2023.
\newblock \href {https://openreview.net/forum?id=jh7wH2AzKK} {Augmented {Language} {Models}: a {Survey}}.
\newblock \emph{Transactions on Machine Learning Research}.

\bibitem[{OpenAI(2024{\natexlab{a}})}]{openai_gpt-4_2024}
OpenAI. 2024{\natexlab{a}}.
\newblock \href {https://doi.org/10.48550/arXiv.2303.08774} {{GPT}-4 {Technical} {Report}}.
\newblock \emph{arXiv preprint}.
\newblock ArXiv:2303.08774 [cs].

\bibitem[{OpenAI(2024{\natexlab{b}})}]{openai_gpt-4o_2024}
OpenAI. 2024{\natexlab{b}}.
\newblock \href {https://doi.org/10.48550/arXiv.2410.21276} {{GPT}-4o {System} {Card}}.
\newblock \emph{arXiv preprint}.
\newblock ArXiv:2410.21276 [cs].

\bibitem[{Park et~al.(2024)Park, Yoon, Park, Lee, Jeong, and Kang}]{chroknowledge_2024}
Yein Park, Chanwoong Yoon, Jungwoo Park, Donghyeon Lee, Minbyul Jeong, and Jaewoo Kang. 2024.
\newblock \href {https://doi.org/10.48550/arXiv.2410.09870} {{ChroKnowledge}: {Unveiling} {Chronological} {Knowledge} of {Language} {Models} in {Multiple} {Domains}}.
\newblock \emph{arXiv preprint}.
\newblock ArXiv:2410.09870 [cs].

\bibitem[{Pavlick(2023)}]{pavlick_symbols_2023}
Ellie Pavlick. 2023.
\newblock \href {https://doi.org/10.1098/rsta.2022.0041} {Symbols and grounding in large language models}.
\newblock \emph{Philosophical Transactions of the Royal Society A: Mathematical, Physical and Engineering Sciences}, 381(2251).

\bibitem[{Perez et~al.(2021)Perez, Kiela, and Cho}]{perez_true_2021}
Ethan Perez, Douwe Kiela, and Kyunghyun Cho. 2021.
\newblock \href {https://openreview.net/forum?id=ShnM-rRh4T} {True {Few}-{Shot} {Learning} with {Language} {Models}}.
\newblock In \emph{Advances in Neural Information Processing Systems}.

\bibitem[{Pérez-Ortiz et~al.(2021{\natexlab{a}})Pérez-Ortiz, Rivasplata, Parrado-Hernandez, Guedj, and Shawe-Taylor}]{perez-ortiz_progress_2021}
Maria Pérez-Ortiz, Omar Rivasplata, Emilio Parrado-Hernandez, Benjamin Guedj, and John Shawe-Taylor. 2021{\natexlab{a}}.
\newblock \href {https://doi.org/10.48550/arXiv.2111.07737} {Progress in {Self}-{Certified} {Neural} {Networks}}.
\newblock \emph{arXiv preprint}.
\newblock ArXiv:2111.07737 [cs].

\bibitem[{Pérez-Ortiz et~al.(2021{\natexlab{b}})Pérez-Ortiz, Rivasplata, Shawe-Taylor, and Szepesvári}]{perez-ortiz_tighter_2021}
María Pérez-Ortiz, Omar Rivasplata, John Shawe-Taylor, and Csaba Szepesvári. 2021{\natexlab{b}}.
\newblock Tighter risk certificates for neural networks.
\newblock \emph{J. Mach. Learn. Res.}, 22(1):227:10326--227:10365.

\bibitem[{Qiu et~al.(2024)Qiu, Zhao, Ziser, Korhonen, Ponti, and Cohen}]{qiu_tempground_2024}
Yifu Qiu, Zheng Zhao, Yftah Ziser, Anna Korhonen, Edoardo Ponti, and Shay Cohen. 2024.
\newblock \href {https://doi.org/10.18653/v1/2024.naacl-long.391} {Are {Large} {Language} {Model} {Temporally} {Grounded}?}
\newblock In \emph{Proceedings of the 2024 {Conference} of the {North} {American} {Chapter} of the {Association} for {Computational} {Linguistics}: {Human} {Language} {Technologies} ({Volume} 1: {Long} {Papers})}, pages 7064--7083, Mexico City, Mexico. Association for Computational Linguistics.

\bibitem[{Reich et~al.(2010)Reich, Green, Kircher, Krause, Patterson, Durand, Viola, Briggs, Stenzel, Johnson, Maricic, Good, Marques-Bonet, Alkan, Fu, Mallick, Li, Meyer, Eichler, Stoneking, Richards, Talamo, Shunkov, Derevianko, Hublin, Kelso, Slatkin, and Pääbo}]{reich_genetic_2010}
David Reich, Richard~E. Green, Martin Kircher, Johannes Krause, Nick Patterson, Eric~Y. Durand, Bence Viola, Adrian~W. Briggs, Udo Stenzel, Philip L.~F. Johnson, Tomislav Maricic, Jeffrey~M. Good, Tomas Marques-Bonet, Can Alkan, Qiaomei Fu, Swapan Mallick, Heng Li, Matthias Meyer, Evan~E. Eichler, Mark Stoneking, Michael Richards, Sahra Talamo, Michael~V. Shunkov, Anatoli~P. Derevianko, Jean-Jacques Hublin, Janet Kelso, Montgomery Slatkin, and Svante Pääbo. 2010.
\newblock \href {https://doi.org/10.1038/nature09710} {Genetic history of an archaic hominin group from {Denisova} {Cave} in {Siberia}}.
\newblock \emph{Nature}, 468(7327):1053--1060.
\newblock Publisher: Nature Publishing Group.

\bibitem[{Reimers and Gurevych(2019)}]{sentence-bert_2019}
Nils Reimers and Iryna Gurevych. 2019.
\newblock \href {https://doi.org/10.18653/v1/D19-1410} {Sentence-{BERT}: {Sentence} {Embeddings} using {Siamese} {BERT}-{Networks}}.
\newblock In \emph{Proceedings of the 2019 {Conference} on {Empirical} {Methods} in {Natural} {Language} {Processing} and the 9th {International} {Joint} {Conference} on {Natural} {Language} {Processing} ({EMNLP}-{IJCNLP})}, pages 3982--3992, Hong Kong, China. Association for Computational Linguistics.

\bibitem[{Song et~al.(2024)Song, Xu, Zhou, and Neubig}]{song_beyondapi_2024}
Yueqi Song, Frank Xu, Shuyan Zhou, and Graham Neubig. 2024.
\newblock \href {https://doi.org/10.48550/arXiv.2410.16464} {Beyond {Browsing}: {API}-{Based} {Web} {Agents}}.
\newblock \emph{arXiv preprint}.
\newblock ArXiv:2410.16464 [cs].

\bibitem[{Strzelecki and Rutecka(2020)}]{strzelecki_direct_2020}
Artur Strzelecki and Paulina Rutecka. 2020.
\newblock \href {https://doi.org/10.1109/ACCESS.2020.2999160} {Direct {Answers} in {Google} {Search} {Results}}.
\newblock \emph{IEEE Access}, 8:103642--103654.

\bibitem[{Tang et~al.(2024)Tang, Zhang, Jin, Yu, Wang, Jin, Zhang, and Du}]{tang_time_2024}
Hua Tang, Chong Zhang, Mingyu Jin, Qinkai Yu, Zhenting Wang, Xiaobo Jin, Yongfeng Zhang, and Mengnan Du. 2024.
\newblock \href {https://doi.org/10.48550/arXiv.2402.10835} {Time {Series} {Forecasting} with {LLMs}: {Understanding} and {Enhancing} {Model} {Capabilities}}.
\newblock \emph{arXiv preprint}.
\newblock ArXiv:2402.10835 [cs].

\bibitem[{Truong et~al.(2019)Truong, Kuhlmann, Bonyadi, Querlioz, Zhou, and Kavehei}]{truong_epileptic_2019}
Nhan~Duy Truong, Levin Kuhlmann, Mohammad~Reza Bonyadi, Damien Querlioz, Luping Zhou, and Omid Kavehei. 2019.
\newblock \href {https://doi.org/10.1109/ACCESS.2019.2944691} {Epileptic {Seizure} {Forecasting} {With} {Generative} {Adversarial} {Networks}}.
\newblock \emph{IEEE Access}, 7:143999--144009.

\bibitem[{Vinuesa et~al.(2024)Vinuesa, Rabault, Azizpour, Bauer, Brunton, Elofsson, Jarlebring, Kjellstrom, Markidis, Marlevi, Cinnella, and Brunton}]{vinuesa_opportunities_2024}
Ricardo Vinuesa, Jean Rabault, Hossein Azizpour, Stefan Bauer, Bingni~W. Brunton, Arne Elofsson, Elias Jarlebring, Hedvig Kjellstrom, Stefano Markidis, David Marlevi, Paola Cinnella, and Steven~L. Brunton. 2024.
\newblock \href {https://doi.org/10.48550/arXiv.2405.04161} {Opportunities for machine learning in scientific discovery}.
\newblock \emph{arXiv preprint}.
\newblock ArXiv:2405.04161 [cs].

\bibitem[{Wachter and Brynjolfsson(2024)}]{wachter_genai_2024}
Robert~M. Wachter and Erik Brynjolfsson. 2024.
\newblock \href {https://doi.org/10.1001/jama.2023.25054} {Will {Generative} {Artificial} {Intelligence} {Deliver} on {Its} {Promise} in {Health} {Care}?}
\newblock \emph{JAMA}, 331(1):65--69.

\bibitem[{Wallat et~al.(2024)Wallat, Jatowt, and Anand}]{wallat_temporal_2024}
Jonas Wallat, Adam Jatowt, and Avishek Anand. 2024.
\newblock \href {https://doi.org/10.1145/3616855.3635818} {Temporal {Blind} {Spots} in {Large} {Language} {Models}}.
\newblock In \emph{Proceedings of the 17th {ACM} {International} {Conference} on {Web} {Search} and {Data} {Mining}}, {WSDM} '24, pages 683--692, New York, NY, USA. Association for Computing Machinery.

\bibitem[{Wang et~al.(2024)Wang, Cheng, Zhu, Fried, and Neubig}]{wang_tools_2024}
Zhiruo Wang, Zhoujun Cheng, Hao Zhu, Daniel Fried, and Graham Neubig. 2024.
\newblock \href {https://openreview.net/forum?id=Xh1B90iBSR} {What {Are} {Tools} {Anyway}? {A} {Survey} from the {Language} {Model} {Perspective}}.
\newblock In \emph{First Conference on Language Modeling}.

\bibitem[{Wei et~al.(2022)Wei, Wang, Schuurmans, Bosma, Ichter, Xia, Chi, Le, and Zhou}]{wei_cot_2022}
Jason Wei, Xuezhi Wang, Dale Schuurmans, Maarten Bosma, Brian Ichter, Fei Xia, Ed~H. Chi, Quoc~V. Le, and Denny Zhou. 2022.
\newblock \href {https://openreview.net/forum?id=_VjQlMeSB_J} {Chain-of-{Thought} {Prompting} {Elicits} {Reasoning} in {Large} {Language} {Models}}.
\newblock In \emph{Advances in Neural Information Processing Systems}.

\bibitem[{Wornow et~al.(2024)Wornow, Narayan, Opsahl-Ong, McIntyre, Shah, and Ré}]{wornow_automating_2024}
Michael Wornow, Avanika Narayan, Krista Opsahl-Ong, Quinn McIntyre, Nigam Shah, and Christopher Ré. 2024.
\newblock \href {https://doi.org/10.14778/3681954.3681964} {Automating the {Enterprise} with {Foundation} {Models}}.
\newblock \emph{Proc. VLDB Endow.}, 17(11):2805--2812.

\bibitem[{Wuestman et~al.(2020)Wuestman, Hoekman, and Frenken}]{wuestman_typology_2020}
Mignon Wuestman, Jarno Hoekman, and Koen Frenken. 2020.
\newblock \href {https://doi.org/10.1162/qss_a_00079} {A typology of scientific breakthroughs}.
\newblock \emph{Quantitative Science Studies}, 1(3):1203--1222.

\bibitem[{Xian et~al.(2025)Xian, Baker, David, Cui, Holmgren, Bauer, Sushil, and Abbasi-Asl}]{xian_robustness_2025}
R.~Patrick Xian, Noah~R. Baker, Tom David, Qiming Cui, A.~Jay Holmgren, Stefan Bauer, Madhumita Sushil, and Reza Abbasi-Asl. 2025.
\newblock \href {https://doi.org/10.48550/arXiv.2502.10374} {Robustness tests for biomedical foundation models should tailor to specification}.
\newblock \emph{arXiv preprint}.
\newblock ArXiv:2502.10374 [cs].

\bibitem[{Xiong et~al.(2024)Xiong, Payani, Kompella, and Fekri}]{xiong_large_2024}
Siheng Xiong, Ali Payani, Ramana Kompella, and Faramarz Fekri. 2024.
\newblock \href {https://doi.org/10.18653/v1/2024.acl-long.563} {Large {Language} {Models} {Can} {Learn} {Temporal} {Reasoning}}.
\newblock In \emph{Proceedings of the 62nd {Annual} {Meeting} of the {Association} for {Computational} {Linguistics} ({Volume} 1: {Long} {Papers})}, pages 10452--10470, Bangkok, Thailand. Association for Computational Linguistics.

\bibitem[{Xu et~al.(2024)Xu, Song, Li, Tang, Jain, Bao, Wang, Zhou, Guo, Cao, Yang, Lu, Martin, Su, Maben, Mehta, Chi, Jang, Xie, Zhou, and Neubig}]{theagentcompany_2024}
Frank~F. Xu, Yufan Song, Boxuan Li, Yuxuan Tang, Kritanjali Jain, Mengxue Bao, Zora~Z. Wang, Xuhui Zhou, Zhitong Guo, Murong Cao, Mingyang Yang, Hao~Yang Lu, Amaad Martin, Zhe Su, Leander Maben, Raj Mehta, Wayne Chi, Lawrence Jang, Yiqing Xie, Shuyan Zhou, and Graham Neubig. 2024.
\newblock \href {https://doi.org/10.48550/arXiv.2412.14161} {{TheAgentCompany}: {Benchmarking} {LLM} {Agents} on {Consequential} {Real} {World} {Tasks}}.
\newblock \emph{arXiv preprint}.
\newblock ArXiv:2412.14161 [cs].

\bibitem[{Yao et~al.(2023)Yao, Zhao, Yu, Du, Shafran, Narasimhan, and Cao}]{yao_react_2022}
Shunyu Yao, Jeffrey Zhao, Dian Yu, Nan Du, Izhak Shafran, Karthik~R. Narasimhan, and Yuan Cao. 2023.
\newblock \href {https://openreview.net/forum?id=WE_vluYUL-X} {{ReAct}: {Synergizing} {Reasoning} and {Acting} in {Language} {Models}}.
\newblock In \emph{The Eleventh International Conference on Learning Representations}.

\bibitem[{Ye et~al.(2024{\natexlab{a}})Ye, Hu, Deng, Huang, Ma, Zhu, and Wang}]{ye_mirai_2024}
Chenchen Ye, Ziniu Hu, Yihe Deng, Zijie Huang, Mingyu~Derek Ma, Yanqiao Zhu, and Wei Wang. 2024{\natexlab{a}}.
\newblock \href {https://doi.org/10.48550/arXiv.2407.01231} {{MIRAI}: {Evaluating} {LLM} {Agents} for {Event} {Forecasting}}.
\newblock \emph{arXiv preprint}.
\newblock ArXiv:2407.01231 [cs].

\bibitem[{Ye et~al.(2024{\natexlab{b}})Ye, Li, Li, Huang, Gao, Wu, Zhang, Gui, and Huang}]{ye_toolsword_2024}
Junjie Ye, Sixian Li, Guanyu Li, Caishuang Huang, Songyang Gao, Yilong Wu, Qi~Zhang, Tao Gui, and Xuanjing Huang. 2024{\natexlab{b}}.
\newblock \href {https://doi.org/10.18653/v1/2024.acl-long.119} {{ToolSword}: {Unveiling} {Safety} {Issues} of {Large} {Language} {Models} in {Tool} {Learning} {Across} {Three} {Stages}}.
\newblock In \emph{Proceedings of the 62nd {Annual} {Meeting} of the {Association} for {Computational} {Linguistics} ({Volume} 1: {Long} {Papers})}, pages 2181--2211, Bangkok, Thailand. Association for Computational Linguistics.

\bibitem[{Zhang et~al.(2009)Zhang, Chang, Zheng, Metzler, and Nie}]{zhang_tempsearch_2009}
Ruiqiang Zhang, Yi~Chang, Zhaohui Zheng, Donald Metzler, and Jian-yun Nie. 2009.
\newblock \href {https://aclanthology.org/N09-2042/} {Search {Engine} {Adaptation} by {Feedback} {Control} {Adjustment} for {Time}-sensitive {Query}}.
\newblock In \emph{Proceedings of {Human} {Language} {Technologies}: {The} 2009 {Annual} {Conference} of the {North} {American} {Chapter} of the {Association} for {Computational} {Linguistics}, {Companion} {Volume}: {Short} {Papers}}, pages 165--168, Boulder, Colorado. Association for Computational Linguistics.

\bibitem[{Zhao et~al.(2024)Zhao, Brumbaugh, Wang, Hajishirzi, and Smith}]{zhao_stc_2024}
Bowen Zhao, Zander Brumbaugh, Yizhong Wang, Hannaneh Hajishirzi, and Noah Smith. 2024.
\newblock \href {https://doi.org/10.18653/v1/2024.findings-acl.892} {Set the {Clock}: {Temporal} {Alignment} of {Pretrained} {Language} {Models}}.
\newblock In \emph{Findings of the {Association} for {Computational} {Linguistics}: {ACL} 2024}, pages 15015--15040, Bangkok, Thailand. Association for Computational Linguistics.

\bibitem[{Zhou et~al.(2024{\natexlab{a}})Zhou, Xu, Zhu, Zhou, Lo, Sridhar, Cheng, Ou, Bisk, Fried, Alon, and Neubig}]{webarena_2024}
Shuyan Zhou, Frank~F. Xu, Hao Zhu, Xuhui Zhou, Robert Lo, Abishek Sridhar, Xianyi Cheng, Tianyue Ou, Yonatan Bisk, Daniel Fried, Uri Alon, and Graham Neubig. 2024{\natexlab{a}}.
\newblock \href {https://openreview.net/forum?id=oKn9c6ytLx} {{WebArena}: {A} {Realistic} {Web} {Environment} for {Building} {Autonomous} {Agents}}.
\newblock In \emph{The Twelfth International Conference on Learning Representations}.

\bibitem[{Zhou et~al.(2024{\natexlab{b}})Zhou, Liu, Srivastava, Mei, and Tan}]{zhou_hypothesis_2024}
Yangqiaoyu Zhou, Haokun Liu, Tejes Srivastava, Hongyuan Mei, and Chenhao Tan. 2024{\natexlab{b}}.
\newblock \href {https://doi.org/10.18653/v1/2024.nlp4science-1.10} {Hypothesis {Generation} with {Large} {Language} {Models}}.
\newblock In \emph{Proceedings of the 1st {Workshop} on {NLP} for {Science} ({NLP4Science})}, pages 117--139, Miami, FL, USA. Association for Computational Linguistics.

\end{thebibliography}

\end{document}